\documentclass[conference]{IEEEtran}
\IEEEoverridecommandlockouts
\usepackage{hyperref}
\hypersetup{
    colorlinks=true,
    linkcolor=blue,
    filecolor=magenta,      
    urlcolor=cyan,
}
\usepackage{cite}
\usepackage{amsmath,amssymb,amsfonts}
\usepackage{algorithmic}
\usepackage{graphicx}
\usepackage{textcomp}
\usepackage{xcolor}
\def\BibTeX{{\rm B\kern-.05em{\sc i\kern-.025em b}\kern-.08em
    T\kern-.1667em\lower.7ex\hbox{E}\kern-.125emX}}

\newcommand\blfootnote[1]{%
  \begingroup
  \renewcommand\thefootnote{}\footnote{#1}%
  \addtocounter{footnote}{-1}%
  \endgroup
}
\makeatletter
\def\footnoterule{\kern-3\p@
  \hrule \@width 1.4in \kern 2.6\p@} 
\makeatother

\begin{document}

\title{Exploring Temporal Context and Human Movement Dynamics
for Online Action Detection in Videos}

\author{\IEEEauthorblockN{Vasiliki I. Vasileiou, Nikolaos Kardaris and Petros Maragos}
\IEEEauthorblockA{School of E.C.E., National Technical University of Athens, Athens 15773, Greece}
\IEEEauthorblockA{\tt\small silavassiliou2@gmail.com, nkardaris@mail.ntua.gr, maragos@cs.ntua.gr}}
\maketitle
\begin{abstract}
Nowadays, the interaction between humans and robots is constantly expanding, requiring more and more human motion recognition applications to operate in real time. However, most works on temporal action detection and recognition perform these tasks in offline manner, i.e.  temporally segmented videos are classified as a whole. In this paper, based on the recently proposed framework of Temporal Recurrent Networks, we explore how temporal context and human movement dynamics can be effectively employed for online action detection. Our approach uses various state-of-the-art architectures and appropriately combines the extracted features in order to improve action detection. We evaluate our method on a challenging but widely used dataset for temporal action localization, THUMOS'14. Our experiments show significant improvement over the baseline method, achieving state-of-the art results on THUMOS'14.
\end{abstract}


\section{Introduction}
\blfootnote{The work of  N. Kardaris \& P. Maragos  has been co-financed by the EU and Greek national funds through the Operational Program Competitiveness, Entrepreneurship and Innovation, under the call RESEARCH CREATE INNOVATE (iWalk, T1EDK- 01248).}
Human Action Recognition (HAR) is one of the most prominent tasks in the field of computer vision, with various applications in robotics \cite{b35, b36},\cite{b37}, data retrieval \cite{b42, b43}, healthcare \cite{b38, b39} etc. Most works deal with action recognition in an offline setting, i.e. the temporal boundaries of an action in a video are known. However, online applications such as on autonomous cars or assistive robotics require recognition capabilities in a continuous video stream, where the starting and ending points of an action have to be estimated on-the-fly, e.g. in order to avoid a car crash \cite{b40} or even a human fall \cite{b41}. \par
Therefore, the difference between the two approaches lies in the time of the decision. Specifically, in the case of offline recognition the action will be observed in its entirety, whereas in online recognition, the decision will have to be taken before the action is completed. Many methods have been developed for the recognition of human action with remarkable results, however they operate under the condition that recognition is made once the action is completed and the network has been fed with all the necessary information. This condition cannot be however satisfied in an online setting, where only present and past information can be used, making the generalisation of offline methods problematic. \par
The recently proposed Temporal Recurrent Networks (TRN)~\cite{b1} introduced a way to bypass the lack of future information by using recurrent networks to predict features that correspond to future frames. TRN processes videos sequentially and for each frame it combines past, present and predicted future information to extract action class probabilities.
\begin{figure}[tbp]
\centerline{\includegraphics[width=0.45\textwidth]{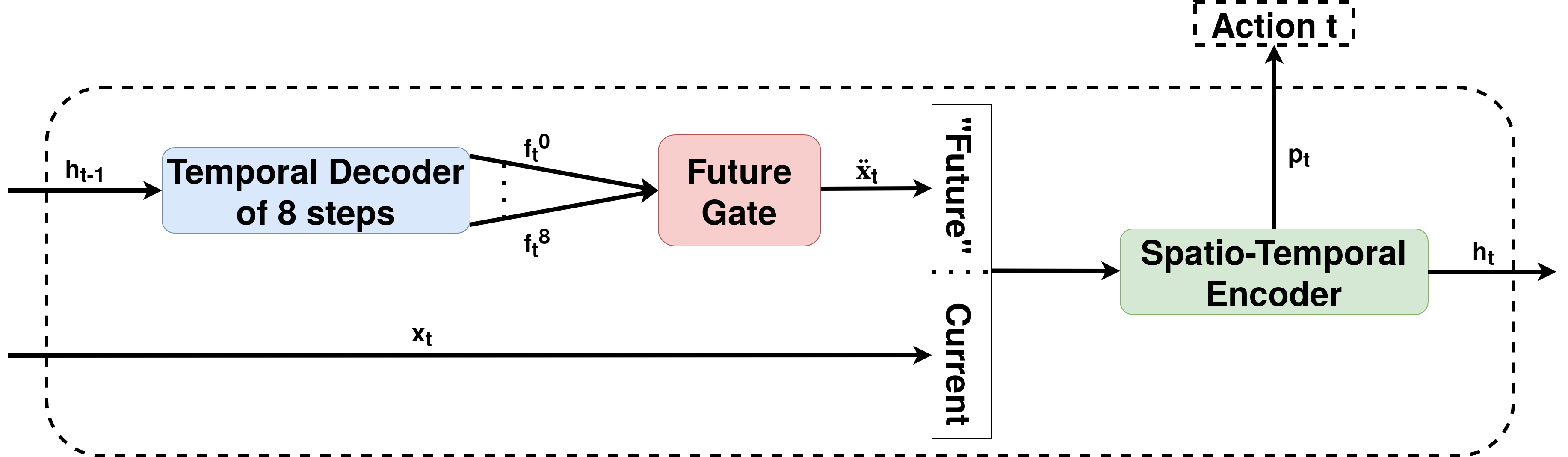}}
\caption{The TRN Cell Architecture as described in \cite{b1}.}
\label{fig1}
\end{figure}
\begin{figure}[tbp]
\centerline{\includegraphics[width=0.46\textwidth]{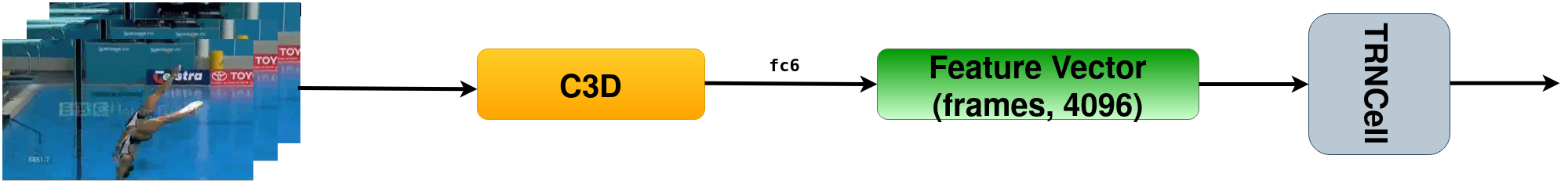}}
\caption{The Pipeline of the One-Stream model which is fed directly by the extracted C3D features.}
\label{fig2}
\end{figure}
Inspired by TRN, we explore different ways to enhance temporal action detection. Our contribution is two-fold:
\begin{itemize}
\item We employ 3D convolutional networks (I3D \cite{b5} \& C3D \cite{b4}) both in action anticipation and prediction. In this way, information about the temporal context of each frame is encoded both by the recurrent network and in terms of visual features.
\item We incorporate human pose in our framework, postulating that the dynamics of the human movement provide valuable information about the temporal boundaries of an action.
\end{itemize}
Two-stream features are introduced as an input to our model which are interpreted as appearance and motion features. Some combinations of the above mentioned extracted features have therefore been selected in order to increase the efficiency of our model. Experiments on the THUMOS14 \cite{b8} public dataset, show significant improvement over the baseline both in action recognition and anticipation. \par
The rest of this paper is organized as follows: Section II provides a review of the literature work in the field. The methodology applied throughout our experiments is analyzed in section III. Section IV  deals with the experimental setup, the dataset and the evaluation methods we utilize, whereas in Section V we discuss the results of our experiments. Finally, in Section VI we present our conclusions and propose directions for further research. \par

\section{Related Work}

\subsection{Human Action Recognition}
Action recognition in videos has stimulated the interest of the research community for years. Traditional methods, utilizing feature descriptors paved the way to more complex neural networks, being proposed nowadays. \par

Early works on action recognition extracted hand-crafted features, such as Histogram of Oriented Gradients (HOG) / Histogram of Optical Flow \cite{b9}, extended Speeded-Up Robust Feature \cite{b11}, Dense Trajectories \cite{b12} and encoded each video using Bag-Of-Words, Fisher vector and other orderless representations. Support Vector Machines were typically used to classify those features. \par

Although such methods have shown promising results \cite{b44}, the abundance of visual data and advances in hardware design led the research community to embrace neural network models. Convolutional Neural Networks (CNN) in particular, have provided significant improvements in image recognition tasks\cite{b18}, something that intuitively bolstered their application to videos and action recognition. Initially, Donahue et al. in \cite{b13} proposed to add a recurrent layer to the CNN to encode state and capture temporal sequence and long-term dependencies. Due to 2D features limitation of 2D CNNs, Ji et al. developed a 3D CNN model \cite{b15} that extracts features from both spatial and temporal dimensions through 3D convolutions, thereby capturing the motion information encoded in multiple adjacent frames. Based on this work, Tran et al. created an efficient descriptor -C3D \cite{b16}- which can be used as a pre-trained feature extractor for other video analysis tasks. In \cite{b14}, a two-stream network, introduced by Simonyan and Zisserman, analyzes spatiotemporal features via RGB images and optical flow. The 3D-fused extension \cite{b19} of the previous model introduces a better performance by fusing spatial and flow streams after the last convolutional layer. Finally, Carreira et al. combined the above models into a new one -I3D \cite{b5}- aiming to very deep, naturally spatio-temporal classifiers. \par
On the other hand Noori et al. based on the fact that skeleton based action recognition can avoid explicitly model the dynamics of actions, propose in \cite{b20} the use of OpenPose \cite{b21} and Recurrent Neural Networks (RNNs) \cite{b22},\cite{b23} to recognize the activities.

\subsection{Offline Action Detection}
Regarding the variant of offline recognition, the sample videos are entirely known a-priori, so the task is to estimate the starting and ending timestamp of each action. This type of problem has the advantages of the whole frame sequence knowledge and the lack of processing time limitation. 
\begin{figure}[htbp]
\centerline{\includegraphics[width=0.50\textwidth]{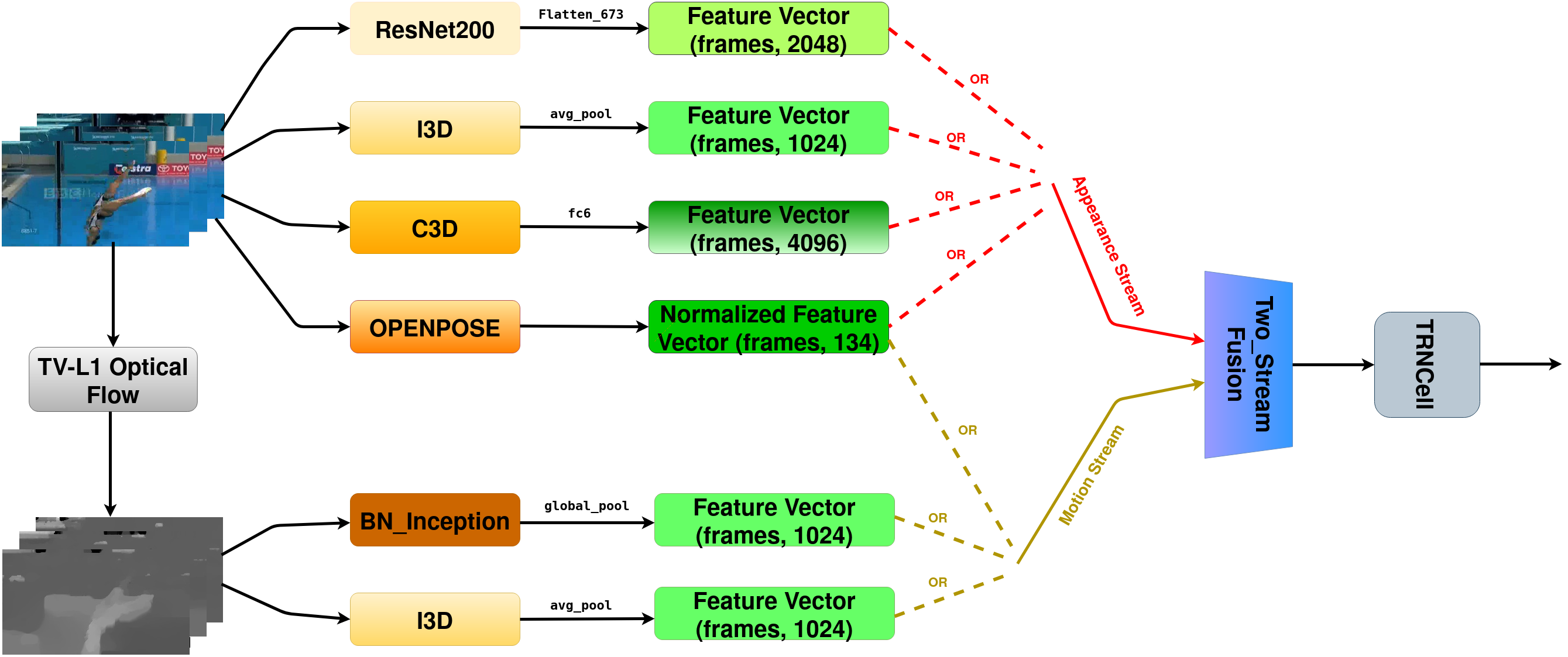}}
\caption{The pipelines of the eleven possible two-stream models, of which the six most effective combinations have been selected. In particular, the selected are: i)ResNet-200 - BN-Inception, ii)I3D (RGB) - I3D (Flow), iii)C3D - OpenPose, iv)ResNet-200 - OpenPose, v)I3D(RGB) - OpenPose, vi)OpenPose - BN-Inception}
\label{fig3}
\end{figure}
Early works \cite{b24},\cite{b25, b26} used sliding windows, where each window is considered as an action candidate, subject to classification. Escorcia et al. \cite{b27} exploit Long Short-Term Memory (LSTM) cells to encode a video sequence as a set of discrete states in order to demonstrate proposal scores, while Gao et al. proposed the TURN TAP \cite{b28} where a long untrimmed video is decomposed into video units, which are then reused as basic building blocks of temporal proposals. Furthermore, Shou et al. \cite{b29} introduced the CDC network, which makes dense per-frame predictions through downsampling in space and upsampling to localize the temporal boundaries. Finally, Long et al. presented GTANs \cite{b30}, which, in contrast to the previous methods, leverage the temporal structure in an one-stage action localization framework.

\subsection{Online Action Detection}
Online Action Detection was initially proposed by De Geest et al. \cite{b31}, who also created the TVSeries dataset for the same purpose. The same research group later proposed a two-stream LSTM model \cite{b32}, focusing both on the interpretation of the frames and on the temporal dependencies between actions. The RED network \cite{b33}, created for action anticipation, can also be used for action detection, if the anticipation time is set to zero. The main idea of this work is the prediction of feature actions using a CNN for feature extraction in combination with a LSTM and a reinforcement loss. Generative Adversarial Networks (GANs) have been also proposed by Shou et al. \cite{b34} to predict starting time precisely. \par
Temporal Recurrent Network (TRN) \cite{b1} proposed by Xu et al. is a novel method which uses the predicted future information to enrich the online action detection accuracy. Based on this work we explore different feature extraction and fusion methods to better estimate information about the temporal dynamics of actions and therefore to achieve higher scores both in anticipation and recognition task.

\section{Methodology}
To address the problem of online action recognition we propose a framework consisting of two main components, one that explores the temporal context of videos, and one that corresponds to the TRN cell proposed in \cite{b1}.
\subsection{TRN Cell}
We first provide a brief description of the TRN cell, which is crucial for the intuition of our method. In particular, the central idea behind TRN is to anticipate future frames' features and aggregate it with past and present information to properly categorize the action. As shown in Fig.~\ref{fig1}, it consists of the temporal decoder, the future gate and the spatiotemporal encoder. Both the temporal decoder and the spatial encoder are LSTM units, where the former accepts serial input vectors and exports the predicted future information and the corresponding hidden states. The second unit sequentially receives the concatenated input and future vectors as well as the hidden state and estimates a probability distribution for each action.
\subsection{Exploiting Contextual Information}
Inspired by the two-stream model used in \cite{b1}, we experimented by extracting I3D features, which are low-level spatial features. Those are computed using an I3D network that is pre-trained on Kinetics \cite{b5}, to improve the ability of the model to generalize and avoid overfitting. We extract the $2\times$($1024$-dimensional) frame-level features from the last global average polling layer. The two-stream features - appearance and motion - are concatenated and fed to a Linear layer with a ReLU activation. Then, the fused information enters the TRN cell, as shown in Fig.~\ref{fig3}.\par
In addition we experimented with C3D features, being a very generic video feature representation \cite{b4}. This is because 3D convolutional modules can extract both spatial and temporal components, as opposed to the ResNet that is utilized by the original method \cite{b1} but limited to appearance representation. The aforementioned one-stream features are computed using C3D network pre-trained on Sports 1M, something that reduces the need to fine-tune. We extract 4096-dimensional features per frame from the fc6-layer and we insert them directly to the TRN cell, as shown in Fig.~\ref{fig2}. \par
Skeleton joint coordinates are of high precision and can accurately represent the temporal dynamics of actions\cite{b45}, so we experimented with 2D skeletons extracted from OpenPose \cite{b6}, over the baseline RBG and Optical Flow features. Thus, 134-dimensional vectors per frame were created. The human pose consists of 25 keypoints for pose/foot estimation and $2\times21$ keypoints for hand estimation. Since 2D models are used, each keypoint consists of two spatial variables, its coordinates and a confidence parameter. 
We first normalize the features we get from OpenPose to address different camera setups. Specifically, we define the middle of the pelvis as the center of our coordinates and normalize with respect to the distance between the pelvis and the shoulders (average height). In the case of multiple actors in a frame the one whose coordinates have the highest confidence score is used. \par
Since the extracted skeleton features are primarily motion features \cite{b45}, we added another stream to the C3D model. 
Specifically, we arranged the C3D features in the appearance stream and the pose features in the motion stream. This framework shows indeed an improvement in performance, compared to the corresponding one-stream model. 
\begin{figure}[htbp]
\centerline{\includegraphics[width=0.50\textwidth]{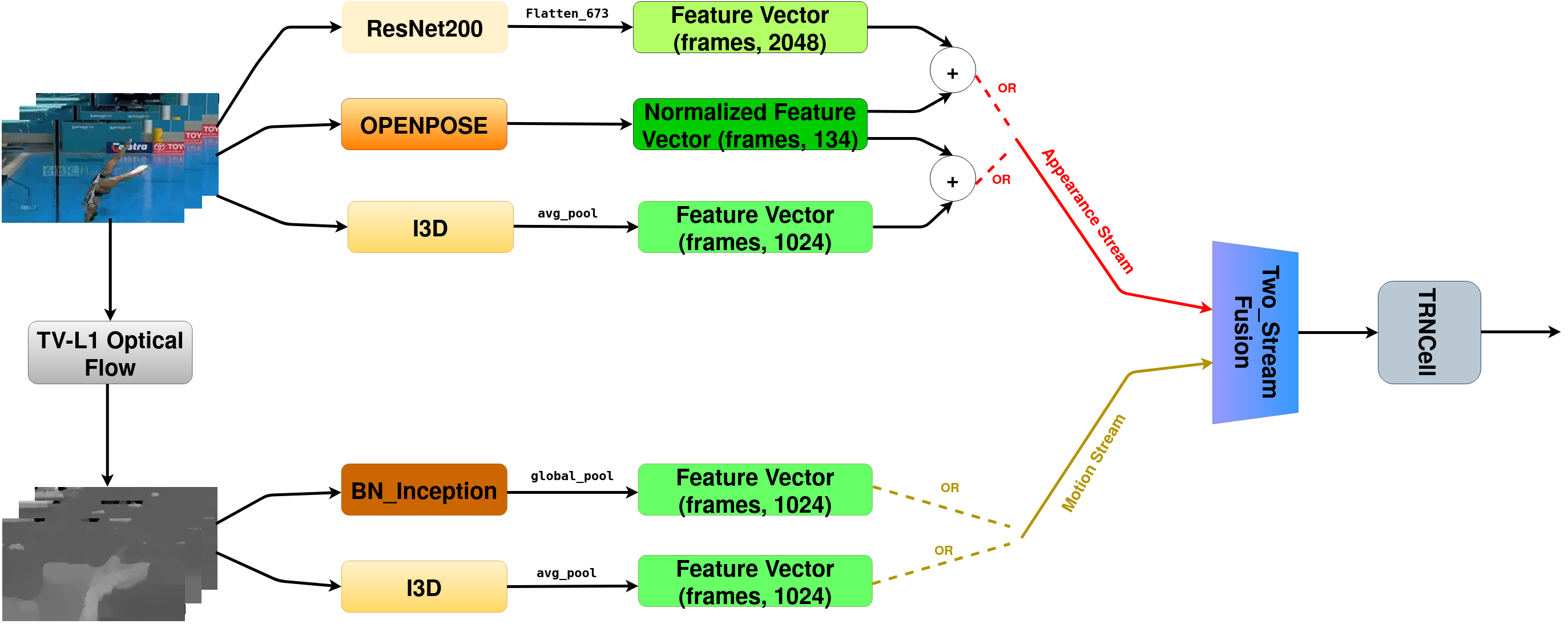}}
\caption{The pipelines of the four possible two-stream fused models, of which the selected are: i) ResNet-200 concatenated with OpenPose - BN-Inception, ii) I3D (RGB) concatenated with Openpose - I3D (Flow).}
\label{fig4}
\end{figure}
Motivated by these results we apply the same framework to the I3D model: we arrange the I3D RGB data in the appearance stream and the OpenPose data in the motion stream.\par
Although the sequences of skeleton features sufficiently represent the temporal dynamics, the appearance and scene information is still missing. Based on the previous claim and on the feedback from our experiments so far, we attempted to combine each of our two-stream models - baseline and I3D - with the information from the skeleton. Specifically, we fused the RGB data with the OpenPose data and created a fused two-stream model as shown in Fig.~\ref{fig4}.

\section{Experimental Setup}
For the evaluation of our model we used the THUMOS'14 dataset \cite{b46} as it contains long and untrimmed videos from various sports events, which are annotated with 20 actions. Its training set however contains only trimmed videos that cannot be used for the task of temporal localization. As a result, based on the previous work \cite{b33}, we train our model on the validation set (200 untrimmed videos) and validate it on the test set (213 untrimmed videos). \par

All experiments were preformed on Nvidia GeForce RTX 2080 Ti GPUs. Adam optimizer was used for the training session \cite{b47} with learning rate and weight decay parameters set to $5\times10^{-4}$. Due to GPU memory limitations, the batch size was set to 2 and the input sequence length was set to 64, whereas we included 8 decoder steps. To permit fair comparisons against the original method \cite{b1}, we performed in-house testing for the baseline TRN with the previous settings. \par

As for video preprocessing, we extracted video frames at 30 fps and experimented with video chunk sizes of 6 and 16, in line with the examined set of experiments. The TV-L1 Optical Flow \cite{b48} algorithm was used to extract the optical flow frames through the Dense-Flow tool. Finally, we report per-frame Mean Average Precision (mAP) for evaluation. To provide more insight, we also report mAP for anticipation times ranging from 0.25 to 2 sec.

\begin{figure*}[ht]
\centerline{\includegraphics[width=0.8\textwidth, height=3cm]{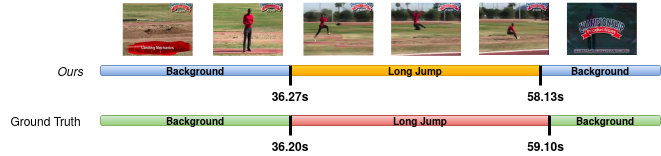}}
\caption{Visualization of our best method - I3D}
\label{fig5}
\end{figure*}

\section{Results}
\begin{table*}[ht]
\caption{Chunk Size 6 Experiments, Using ResNet-200, BN-Inception and OpenPose}
\begin{center}
\resizebox{0.88\textwidth}{0.0410537\textheight}{
\begin{tabular}{|c|c|c|c|c|c|c|c|c|c|c|c|}
\hline
\textbf{Method}&\textbf{Features}&\textbf{Encoder}&\multicolumn{9}{|c|}{\textbf{Decoder - Time predicted into the future (seconds)}} \\
\cline{4-12} 
 & \textbf{Chunk size = 6 frames}& & \textbf{\textit{0.25s}}& \textbf{\textit{0.50s}}& \textbf{\textit{0.75s}}& \textbf{\textit{1.00s}} & \textbf{\textit{1.25s}}& \textbf{\textit{1.50s}}& \textbf{\textit{1.75s}}& \textbf{\textit{2.00s}}& \textbf{\textit{Avg}}\\
\hline
\hline
Baseline& RGB -- Flow& 25.93& \textbf{26.15}& 25.89& 25.79& 25.73& 25.66& \textbf{25.68}& \textbf{25.66}& \textbf{25.57}& \textbf{25.77}\\
\hline
Ours& \{RGB + OpenPose\} -- Flow& 24.25& 23.11& 25.63& \textbf{26.72}& 26.18& 25.57& 24.94& 24.40& 23.94& 25.06\\
\hline
Ours& RGB -- OpenPose& \textbf{37.57}& 25.54& \textbf{25.93}& 26.44& \textbf{26.60}& \textbf{26.28}& 25.57& 24.75& 24.00& 25.64\\
\hline
Ours& OpenPose -- Flow& 36.30& 21.77& 22.59& 23.57& 23.19& 22.28& 21.30& 20.49& 19.83& 21.88\\
\hline
\end{tabular}}
\label{tab1}
\end{center}
\end{table*}
\begin{table*}[ht]
\caption{Chunk Size 16 Experiments, Using C3D and Openpose}
\begin{center}
\resizebox{0.88\textwidth}{0.03\textheight}{
\begin{tabular}{|c|c|c|c|c|c|c|c|c|c|c|c|}
\hline
\textbf{Method}&\textbf{Features}&\textbf{Encoder}&\multicolumn{9}{|c|}{\textbf{Decoder - Time predicted into the future (seconds)}} \\
\cline{4-12} 
 & \textbf{Chunk size = 16 frames}& &\textbf{\textit{0.25s}}& \textbf{\textit{0.50s}}& \textbf{\textit{0.75s}}& \textbf{\textit{1.00s}} & \textbf{\textit{1.25s}}& \textbf{\textit{1.50s}}& \textbf{\textit{1.75s}}& \textbf{\textit{2.00s}}& \textbf{\textit{Avg}}\\
\hline
\hline
Ours&C3D (One-Stream)& 35.43& \textbf{34.34}& \textbf{31.05}& 28.22& 26.46& 25.37& \textbf{24.75}& \textbf{24.39}& \textbf{24.22}& \textbf{27.35}\\
\hline
Ours&\{C3D (RBG)\} -- OpenPose& \textbf{36.44}& 32.98& 30.56& \textbf{28.37}& \textbf{26.61}& \textbf{25.38}& 24.54& 23.78& 23.22& 26.93\\
\hline
\end{tabular}}
\label{tab2}
\end{center}
\end{table*}
\begin{table*}[ht]
\caption{Chunk Size 16 Experiments, Using I3D and OpenPose}
\begin{center}
\begin{tabular}{|c|c|c|c|c|c|c|c|c|c|c|c|}
\hline
\textbf{Method}&\textbf{Features}&\textbf{Encoder}&\multicolumn{9}{|c|}{\textbf{Decoder - Time predicted into the future (seconds)}} \\
\cline{4-12} 
 & \textbf{Chunk size = 16 frames}& &\textbf{\textit{0.25s}}& \textbf{\textit{0.50s}}& \textbf{\textit{0.75s}}& \textbf{\textit{1.00s}} & \textbf{\textit{1.25s}}& \textbf{\textit{1.50s}}& \textbf{\textit{1.75s}}& \textbf{\textit{2.00s}}& \textbf{\textit{Avg}}\\
\hline
\hline
\textbf{Ours}& \textbf{I3D}& \textbf{55.25}& \textbf{52.57}& \textbf{46.69}& \textbf{41.94}& \textbf{38.39}& \textbf{35.90}& \textbf{34.22}& \textbf{33.00}& \textbf{32.08}& \textbf{39.35}\\
\hline
Ours& \{I3D (RGB) + OpenPose\} -- \{I3D (Flow)\}& 49.21& 46.65& 40.78& 36.42& 33.19& 30.90& 29.42& 28.43& 27.71& 34.19\\
\hline
Ours& \{I3D (RGB)\} -- OpenPose& 47.43& 44.59& 40.08& 36.77& 34.24& 32.37& 31.29& 30.56& 30.06& 35.00\\
\hline
Ours& \{I3D (RGB)\} -- \{I3D (Flow) + OpenPose\}& 44.47& 29.55& 31.92& 29.62& 27.21& 25.63& 24.78& 24.20& 23.68& 27.07\\
\hline
\end{tabular}
\label{tab3}
\end{center}
\end{table*}
The utilized models that we mentioned in Sec.~3 are shown in Fig.~ 2,3,4, while the results we obtain from our experiments are depicted in Tables I, II, III. These are organized based on the methodology used to extract the features. Specifically in Table I, ResNet-200 and BN-Inception were used for RGB and Flow features extraction respectively, while, in Table II and III, we used C3D and I3D respectively. In all three tables, we denote the use of OpenPose features, either by replacing or complementing flow and RGB features, thus creating motion features and appearance features respectively. \par
By inspecting Table I, where chunk size has been set to 6 frames, we observe that the baseline method, using ResNet-200 for RGB information and BN-Inception for flow information, displays the highest accuracy of 25.77\% for the precision task, the prediction accuracy for the classification task however is only 25.93\%. By replacing flow information with OpenPose features we notice that, while the average decoder accuracy decreases slightly to 25.67\%, the encoder accuracy is significantly improved and reaches 37.57\%. However, it is true that the use of OpenPose features to enhance or replace RGB information does not provide any further improvement, either in the anticipation phase or in the process of detection. It is worth mentioning though that, in both cases that we used OpenPose, we observe better performance for the period 0.5s - 1.25s, but also much smaller than the baseline in longer-term predictions, something that leads to a decline of the average accuracy in these cases. \par

Table II shows the results for C3D, giving one-stream features, where the chunk size has been set to 16. We highlight that the use of human pose as motion features, by introducing a second stream in our model, gives a boost compared to the simple C3D in the phase of action detection, from 35.43\% for the former to 36.44\% for the latter, as opposed to action anticipation, for which the accuracy drops to 26.93\%. Additionally, we should note the large discrepancy between the performance of short-term and long-term anticipation, reaching as much as 10\%. By comparing this table to the previous one, we observe that OpenPose shows, as motion information, better anticipation performance, compared to the corresponding simple model, in the interval 0.75s - 1.25s. Moreover, although in C3D models we observe larger anticipation accuracy, the action detection accuracy does not exceed that of the models of Table I. \par

The results of employing I3D as well as its variations are shown in Table III, where the chunk size is set to 16 frames. We notice that both the simple I3D model and its modifications show much better performance, with the simplest I3D model giving the biggest boost and reaching 39.35\% in the anticipation phase and 55.25\% in the detection phase. However, the use of OpenPose in this set of experiments, both as an additional cue to RGB and flow information and as a unique motion information did not offer any improvement. On the contrary, it limited its effectiveness. This divergence is likely due to the substantially strong capacity, offered by I3D flow information.

\section{Conclusion}
In this paper, we propose several ways to improve online action detection, building upon Temporal Recurrent Networks. Our results highlight the value of temporal context and human pose as useful cues for localizing action in time. We demonstrate that most of our models outperform the original TRN method \cite{b1} by a significant margin, even though our baseline results are lower than the original paper's due to the smaller batch size we used, with the best of them (I3D) achieving state-of-the-art results. Specifically, observing the variations of models' behavior in the analysis and detection phase, we believe that the use of different models for anticipation and recognition could benefit the task of online action detection. We plan to pursue this goal in our future work.

\end{document}